\documentclass[wcp]{jmlr}

\usepackage{longtable}

\usepackage{booktabs}

\usepackage{comment}
 
\usepackage{color}
\definecolor{Orange}{rgb}{1,0.5,0}
\definecolor{Red}{rgb}{1,0,0}
\definecolor{Blue}{rgb}{0,0,1}

\newcommand{\BL}[1]{\textsf{\textbf{\textcolor{Red}{\footnotesize [BL: #1]}}}}
\newcommand{\GT}[1]{\textsf{\textbf{\textcolor{Blue}{\footnotesize [GT: #1]}}}}

\usepackage{bm}
\usepackage{amsfonts} 
\usepackage{amssymb} 
\DeclareMathOperator*{\argmax}{arg\,max}

\usepackage{algorithm}
\LinesNumbered

\usepackage{booktabs}
\usepackage{longtable}
\usepackage{array}

\usepackage{url}

\usepackage{graphicx}

\title[Making Classifier Chains Resilient to Class Imbalance]{Making Classifier Chains Resilient to Class Imbalance}

  \author{\Name{Bin Liu} \Email{binliu@csd.auth.gr} \\
  \Name{Grigorios Tsoumakas} \Email{greg@csd.auth.gr}\\
  \addr School of Informatics, Aristotle University of Thessaloniki, Thessaloniki, Greece
 }

\begin{document}

\maketitle

\begin{abstract}
Class imbalance is an intrinsic characteristic of multi-label data. Most of the labels in multi-label data sets are associated with a small number of training examples, much smaller compared to the size of the data set. Class imbalance poses a key challenge that plagues most multi-label learning methods. Ensemble of Classifier Chains (ECC), one of the most prominent multi-label learning methods, is no exception to this rule, as each of the binary models it builds is trained from all positive and negative examples of a label. To make ECC resilient to class imbalance, we first couple it with random undersampling. We then present two extensions  of this basic approach, where we build a varying number of binary models per label and construct chains of different sizes, in order to improve the exploitation of majority examples with approximately the same computational budget. Experimental results on 16 multi-label datasets demonstrate the effectiveness of the proposed approaches in a variety of evaluation metrics.    
\end{abstract}

\section{Introduction}

\begin{figure}[htb]
\begin{center}
\small
\subfigure[]{%
\includegraphics[width=.49\textwidth]{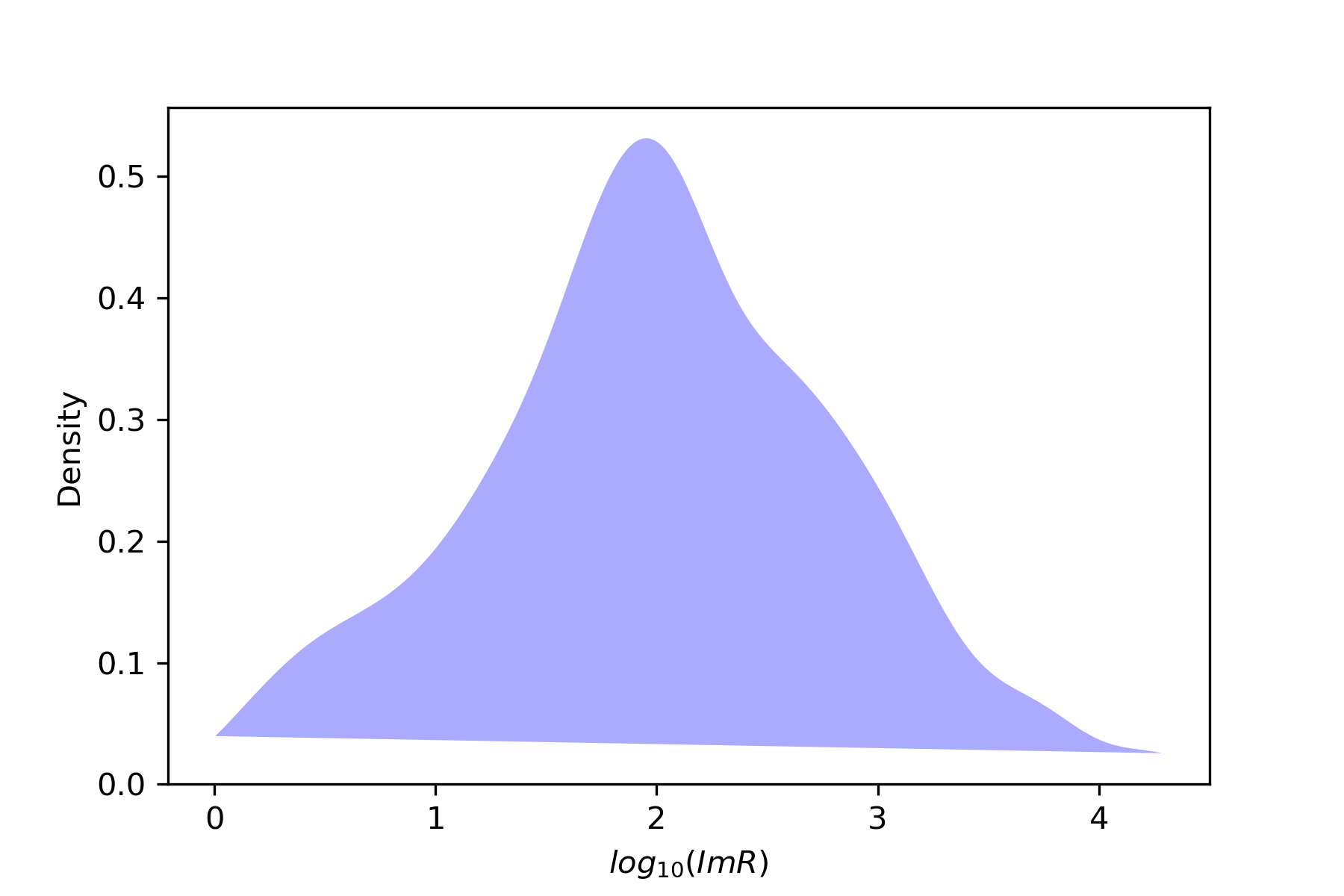}} \label{fig_gkde}
\subfigure[]{%
\includegraphics[width=.49\textwidth]{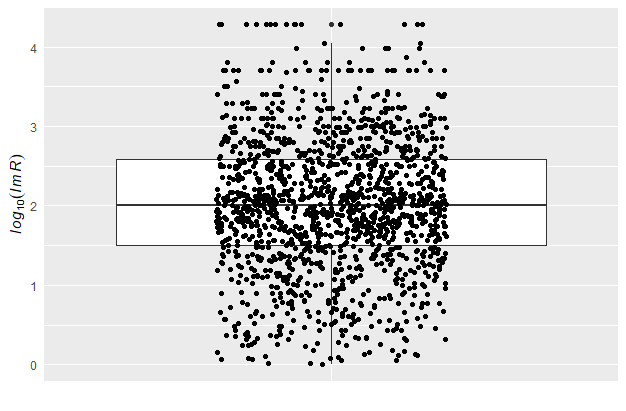}} \label{fig_box}\
\caption{(a) Gaussian kernel density estimation plot and (b) box-plot (values superimposed and jittered) of the imbalance ratios of all labels in the 16 datasets of Table \ref{ta_dataset}.} 
\label{fig_ImR}
\end{center}
\end{figure}

Class imbalance is an intrinsic characteristic of multi-label data. Each training example in a multi-label dataset is typically associated with a small number of labels, much smaller than the total number of labels. This results in a sparse output matrix, where a small total number of positive class values is shared by a much larger number of example-label pairs. Though the distribution of the number of positive class values is not uniform across labels --- in some real-world applications it follows a power law \citep{Rubin2012StatisticalClassification} --- most of the labels are typically associated with a small number of positive class values. The imbalance ratio ($ImR$) of a label is the ratio of the number of examples of the majority class over the number of examples of the minority class. Figure \ref{fig_ImR} (a) shows a density estimation plot and Figure \ref{fig_ImR} (b) a box-plot of the imbalance ratios of all labels in the 16 multi-label datasets of Table \ref{ta_dataset} that are part of our empirical study. We can see indeed that most of the labels are characterized by severe class imbalance.

The starting point of this work is {\em Ensemble of Classifier Chains} (ECC) \citep{Read2011ClassifierClassification}, a popular multi-label learning algorithm with state-of-the-art predictive performance that is also accompanied by a theoretical interpretation based on probability theory \citep{Dembczynski2010BayesChains}. ECC suffers from class imbalance, as each of the binary models it builds is trained from all positive and negative examples of a label. While several approaches have been recently proposed to highlight and address the class imbalance problem in the context of multi-label learning, none of them has considered to build on top of ECC.

To make ECC resilient to class imbalance we contribute a new approach that couples it with random undersampling \citep{Breiman1984ClassificationTrees}. We then present two extensions of this basic approach, in order to improve the exploitation of majority examples with approximately the same computational budget. This is achieved by building a varying number of binary models per label and constructing chains of different sizes. Experimental results on 16 multi-label datasets demonstrate the effectiveness of the proposed approaches in a variety of evaluation metrics.


\section{Our Approach} 
We first introduce the notation used in the rest of the paper and describe the ECC algorithm. Then, we present our approach for making classifier chains resilient to class imbalance along with two extensions that improve the exploitation of majority examples. In the last subsection, we analyze the computational complexity of the proposed methods.

\subsection{Notation}
Let $\mathcal{X}=\mathbb{R}^d$ be a $d$-dimensional input feature space, $L=\{l_1,l_2,...,l_q\}$ a label set containing $q$ labels and $\mathcal{Y}=\{0,1\}^q$ a $q$-dimensional label space. $D=\{(\bm{x}_i,\bm{y}_i)| 1 \leqslant i \leqslant n \}$ is a multi-label training data set containing $n$ instances. Each instance $(\bm{x}_i,\bm{y}_i)$ consists of a feature vector $\bm{x}_i \in \mathcal{X}$ and a label vector $\bm{y}_i \in \mathcal{Y}$, where $y_{ij}$ is the $j$-th element of $\bm{y}_i$ and $y_{ij}=1 (0)$ denotes that $l_j$ is (not) associated with $i$-th instance. 
For label $l_j$, $m_j=\min(|{D}_j^0|,|{D}_j^1|)$ and $M_j=\max(|{D}_j^0|,|{D}_j^1|)$ denote the number of minority and majority class examples respectively, where $D_j^b=\{(\bm{x}_i,\bm{y}_i)|y_{ij}=b, 1 \leqslant i \leqslant n\}$. $ImR_j=M_j/m_j$ is the imbalance ratio of $l_j$. A multi-label method learns a mapping function $h:\mathcal{X} \to \{0,1\}^q$ or $f: \mathcal{X} \to \mathbb{R}^q$ from $D$ that given an unseen instance $\bm{x}$, outputs a label or real-valued vector $\hat{\bm{y}}$ with the predicted labels of or corresponding relevance degrees to $\bm{x}$ respectively.

\subsection{Ensemble of Classifier Chains}
Classifier Chain (CC) is a well-known multi-label learning method that is based on the idea of chaining binary models \citep{Read2011ClassifierClassification}. CC exploits high-order label correlations by sequentially constructing one binary classifier for each label based on a chain (permutation) of the labels $CH$, where $CH_j$ is the index of the label in $L$.
The $j$-th classifier $h_j$ is constructed by the binary dataset whose class is label $l_{CH_j}$ and the feature space of training instances is extended with the values of the previous labels in the chain. Once the classifier chain $\{h_1,...,h_q\}$ is built, the unseen instance $\bm{x}$ is predicted by traversing all classifiers iteratively. The input of $h_j$ is the $\bm{x}$ augmented by predictions of all preceding labels obtained from previous classifiers.


The performance of CC is highly affected by the sequence of the labels within the chain. To relieve the impact of label ordering and make the model more robust, the ECC algorithm constructs $c$ different chains and corresponding CC models \citep{Read2011ClassifierClassification}. To make these models more diverse, each chain is trained on a  different training set $D'$ obtained by 
sampling with replacement ($|D'|=|D|$). The prediction of ECC for a test instance is obtained by combining the predictions of all CCs with a voting strategy. The $j$-th element of relevance degree vector $\hat{\bm{y}}$, denoted by $\hat{y}_j$, is calculated as the number of CCs that predicts $l_j$ as the relevant label of $\bm{x}$ divided by the number of chains $c$.

\subsection{Ensemble of Classifier Chains with Random Undersampling}

To deal with the class imbalance inherent in multi-label data, we firstly propose coupling CC with random undersampling \citep{Breiman1984ClassificationTrees}, in order to balance the class distribution of each binary training set. This leads to the classifier chain with random undersampling approach (CCRU), whose pseudocode is shown in Algorithm \ref{al_CCRUTrain}. 

CCRU builds binary classifiers sequentially according to label sequence $CH$. Random undersampling of majority examples is applied to each binary training set before building the corresponding classifier (line 4). In specific, $M_j-m_j$ majority class examples are randomly removed from each label $l_j$ in order to create a fully balanced training set. 

In the original CC model, the true values of the labels are considered when using them as input features. Recent work found that two alternative approaches lead to better results in the context of multi-target regression chains \citep{Spyromitros-Xioufis2016}: i) using in-sample estimates of the values of these labels by considering the predictions of the corresponding binary models on the training set, ii) using out-of-sample estimates of the values of these labels by considering the cross-validated predictions of the corresponding binary models on the training set. CCRU avoids the second approach because cross-validation would construct training sets that are further deprived of the already small number of minority examples, leading to a deviant distribution of predictions compared to the predictions of the corresponding binary models. In addition, cross-validation is very time consuming. Instead, CCRU follows the first of the above approaches, i.e it considers the predictions of the corresponding binary models on the training set. As only a subset of the majority examples of the training set are used for the training of the corresponding binary model (line 5), CCRU essentially considers a mixture of in-sample and out-of-sample predictions: in-sample for the minority and the equal number of retained majority examples, and out-of-sample for the rest of the majority examples that were removed (lines 7-14). 

Similar to ECC, the Ensemble of Classifier Chains with Random Undersampling (ECCRU) algorithm aggregates several CCRUs that are built upon different label sequences and resampled versions of the original training set.

\begin{algorithm2e}[t]
 \SetKwData{Left}{left}\SetKwData{This}{this}\SetKwData{Up}{up}
 \SetKwFunction{RandomPermute}{RandomPermute}  \SetKwFunction{RandomUnderSample}{RandomUnderSample}
 \SetKwInOut{Input}{input}\SetKwInOut{Output}{output}
 
 \Input{multi-label data set: $D$, sequence of labels: $CH$}
 \Output{CCRU model: $h=\{h_1,...,h_{|CH|}\}$}
 $D_1 \leftarrow \{ (\bm{x}_1,y_{1CH_1}),...,(\bm{x}_{|D|},y_{|D|CH_1})\}$ \;
 $h \leftarrow \emptyset$ \;
\For {$j\leftarrow 1$ \KwTo $|CH|$}{
 	$D^*_j\leftarrow \RandomUnderSample(D_j)$ \tcc*[r]{apply random undersampling to $D_j$}  
    train $h_j$ based on $D^*_j$ \;
    $h \leftarrow h \cup h_j$ \;
    \If{$j<|CH|$}{
        $D_{j+1} \leftarrow \emptyset$ \;
        \ForEach {$(\bm{x},y)$ in $D_j$}{
      		$\hat{y}_{CH_j} \leftarrow h_j(\bm{x}) $ \;
          	$\bm{x}' \leftarrow [x_1,...,x_d,\hat{y}_{CH_1},...,\hat{y}_{CH_{j}}]$  \tcc*[r]{add augmented features}
         	$D_{j+1} \leftarrow D_{j+1} \cup (\bm{x}',y_{CH_{j+1}})$ \;
       	}                     
     }
  }     
\KwRet{$h=\{h_1,...,h_{|CH|}\}$} \;

 \caption{Training of CCRU}
  \label{al_CCRUTrain}
\end{algorithm2e}

\subsection{Improving the Exploitation of Majority Examples}

In ECCRU, the probability that a majority example of a label is eventually used for training the binary models of that label depends on the number of minority, $m$, and majority, $M$, examples of that label, as well as on the number of chains, $c$. In each chain, sampling of all the training examples with replacement is first performed once, followed by separate samplings of the majority examples of each label without replacement. If we skip the first sampling process for the sake of simplifying the analysis, then the probability that a majority example of a label is selected in at least one of the $c$ chains of ECCRU, denoted as $P$, can be obtained by Equation \ref{sampling_prob}.

\begin{equation}
\label{sampling_prob}
P=1-\left(1-\frac{m}{M}\right)^c
\end{equation}

\begin{figure}[ht]
\begin{center}
\includegraphics[width=0.65\textwidth]{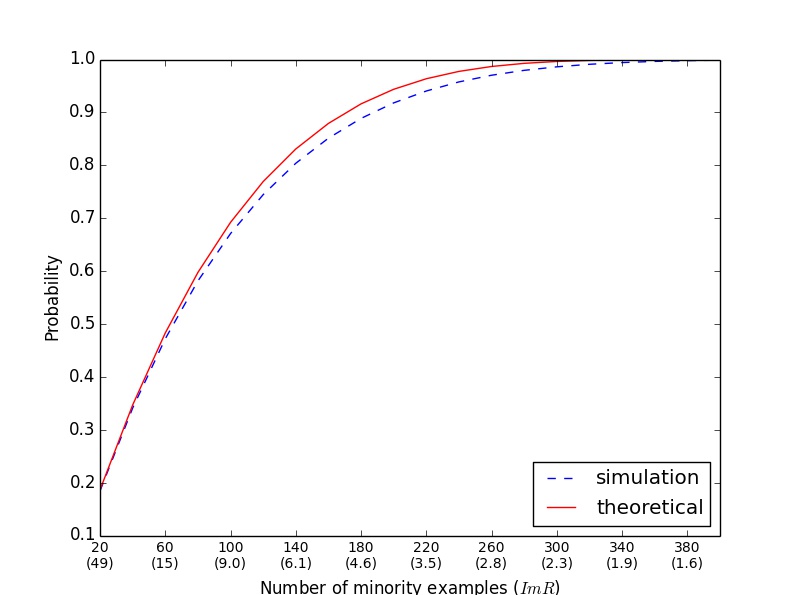}
\caption{The empirically estimated (simulated) and theoretically approximated probability of a majority example of a label being retained for training by at least one of the 10 corresponding models of ECCRU with 10 chains, assuming 1,000 training examples and a number of minority examples varying from 20 to 400 with a step of 20. For the simulation, the sampling process was conducted 10,000 times.}
\label{fig_Pro}
\end{center}
\end{figure}

Figure \ref{fig_Pro} plots Equation \ref{sampling_prob} for 10 chains, 1,000 training examples and varying number of minority samples, as well as the empirical probability in question estimated using 10,000 runs. We notice that for $ImR > 15$ this probability is less than 0.5, with an alternative interpretation being that less than half of the majority examples are eventually used by ECCRU in such a case. As intuitively expected, we see that the higher the $ImR$ of a label, the lower the exploitation of its majority examples.

A straightforward way to increase the exploitation of majority examples when $ImR$ is high is to increase the number of chains. This is also theoretically grounded based on Equation \ref{sampling_prob}. Increasing the number of chains however leads to increased computational cost. We instead consider a variation of our algorithm that improves the exploitation of majority examples without increasing the computational budget. A key observation is that each label contributes a different computational cost to ECCRU, which is proportional to the number of its minority examples, as each corresponding classifier is trained with twice that number of examples. Consider for example a dataset with 100 training examples and 3 labels, each with 10, 20 and 30 minority examples respectively. The classifier of the first, second and third label will be trained with 20, 40 and 60 training examples respectively. 

Our proposal is to redistribute this computational cost by building a different number of classifiers per label, inversely proportional to its number of minority examples. This way we can achieve uniform exploitation of majority examples across labels at the same computational cost. We call this variation of our approach ECCRU2. Continuing the previous example, if we build 10 chains, then the total number of exploited majority examples is 600 (10 times 10+20+30). Our approach divides this computational budget equally across labels, i.e. 200 majority examples per label. We then divide this with the number of minority examples of each label to get the number of classifiers to build for each label, i.e. 20, 10 and $6.\overline{6}$. In general, given $q$ labels and a budget of $c$ chains, the number of classifiers, $c_j$, constructed by ECCRU2 for label $j$ is given by Equation \ref{eq_c_j}. 

\begin{equation}
c_j = \lfloor \frac{c\sum_{k=1}^{q}m_k}{qm_j} \rfloor
\label{eq_c_j}
\end{equation}

To accommodate the fact that the number of classifiers to be constructed differs among labels, ECCRU2 considers partial chains containing an increasingly smaller subset of labels until the minimum size of two labels. Continuing the aforementioned example, ECCRU2 would build 6 chains containing all three labels and 4 chains including the first two labels. 

The pseudo-code of the training process of ECCRU2 is given in Algorithm \ref{al_ECCRU2Train}. Firstly, the number of classifiers trained for each label $c_j$ is calculated according to Equation \eqref{eq_c_j}. To limit the number of classifiers in the case of highly imbalanced labels, we confine $c_j$ to be less than a predefined maximal value $c_{max}$, defined as a multiple of $c$: $c_{max}=c\theta_{max}$  (line 3). In our empirical study we set $c=10$, $\theta_{max}=10$ and therefore $c_{max}=100$.   
Then, in each iteration of building a CCRU model, labels whose corresponding counter $cn_j$ recording the number of classifier needed to be trained is larger than 0 are added into the label set $S$, and only labels collected in $S$ are utilized to generate the label sequence to train the current CCRU model. The loop (in line 6-21) terminates when $c_{max}$ chains have been built or $|S|<2$. The rest parts of the training phase of ECCRU2 are identical to ECCRU. 

The pseudo-code of the testing process of ECCRU2 is given in Algorithm \ref{al_ECCRU2Test}. In ECCRU2, the number of binary classifiers contained in CCRU $h^i$, denoted as $|h^i|$, does not always equal $q$. Hence, a $q$ dimensional vector $\bm{cc}$ is introduced to count the number of binary classifiers for each label, which is used in line 14 to normalize the $\hat{y}_j$, for $j=1,...q$. The rest parts of testing process of ECCRU2 are as in ECC.

One issue in ECCRU2 is that very few classifiers, even just one, can be built in the case of balanced labels with large $m_j$, leading to fewer full-sized chains being built. To address this problem, a variant of ECCRU2 called ECCRU3 is proposed. The only change in ECCRU3 is the addition of a lower bound $c_{min}$ for $c_j$, where $c_{min}=c\theta_{min}$ and $\frac{1}{c} \leqslant \theta_{min} \leqslant 1$ to ensure that $1 \leqslant c_{min} \leqslant c$. Hence, the confined $c_j$ ($cn_j$) is computed as 
$\min \{ \max\{c_j, c\theta_{min}\} , c\theta_{max} \}$ and so at least $c_{min}$ chains containing all of the labels are built. In our empirical study we set $c=10$, $\theta_{min}=0.5$ and therefore $c_{min}=5$. The rest parts of the training and testing process of ECCRU3 are the same with ECCRU2.
\begin{algorithm2e}[t]
\begin{small}
 \SetKwData{Left}{left}\SetKwData{This}{this}\SetKwData{Up}{up}
 \SetKwFunction{RandomPermute}{RandomPermute}  \SetKwFunction{SampleWithReplacement}{SampleWithReplacement}
 \SetKwFunction{TrainCCRU}{TrainCCRU}
 \SetKwInOut{Input}{input}\SetKwInOut{Output}{output}
 
 \Input{multi-label data set: $D$, number of labels: $q$, standard number of chains: $c$, the coefficient of maximal number of chains: $\theta_{max}$}
 \Output{ECCRU2 model: $h=\{h^1,...,h^{c'}\}$}
  \For{$j \leftarrow 1$ \KwTo $q$}{
  	calculate $c_j$ according to \eqref{eq_c_j} \;
    $c_j \leftarrow \min\{c_j,c\theta_{max}\}$\;
    $cn_j \leftarrow c_j$ \tcc*[r]{the number of classifiers needed to be built for each label}
  }
  $c' \leftarrow 0$ \tcc*[r]{the counter to record the number of chains built actually}
  \For {$i \leftarrow 1$ \KwTo $c\theta_{max}$}{
  	$S \leftarrow \emptyset$ \;
    \For {$j \leftarrow 1$ \KwTo $q$}{
    	\If{$cn_j>0$}{
        	$S \leftarrow S \cup j$ \;
            $cn_j \leftarrow cn_j-1$ \;
        }
    }
    \If{$|S|<2$}{
    	\textbf{break}\;
    }
 	$CH^i \leftarrow \RandomPermute (S)$  \tcc*[r]{generate a chain by random permutation}
    $D' \leftarrow \SampleWithReplacement(D)$ \tcc*[r]{sample the $D$ with replacement}
	$h^i \leftarrow \TrainCCRU(D',CH^i)$ \tcc*[r]{train a CCRU according to Algorithm \ref{al_CCRUTrain}}
    $c' \leftarrow c'+1$ \;
  }
  $h\leftarrow \{h^1,...,h^{c'}\}$ \;
\KwRet{$h=\{h^1,...,h^{c'}\}$} \;

 \caption{Training of ECCRU2}
  \label{al_ECCRU2Train}
\end{small}
\end{algorithm2e}

\begin{algorithm2e}[t]
\begin{small}
 \SetKwData{Left}{left}\SetKwData{This}{this}\SetKwData{Up}{up}
 \SetKwFunction{labelInDataset}{labelInDataset}\SetKwFunction{caculateMeanIR}{caculateMeanIR}
 \SetKwInOut{Input}{input}\SetKwInOut{Output}{output} 
 \Input{test instance $\bm{x}$, number of labels: $q$, ECCRU2 model: $h=\{h^1,h^2,...,h^{c'}\}$ 
 }
 \Output{relevance degree vector $\bm{ \hat{y} } $}
  
  $\hat{\bm{y}} \leftarrow \bm{0}$ \; 
  $\bm{cc} \leftarrow \bm{0}$ \tcc*[r]{$\bm{cc}$ is a $q$ dimensional counter}
  \For {$i\leftarrow 1$ \KwTo $c'$}{
     \For {$j\leftarrow 1$ \KwTo $|h^i|$}{
        $k \leftarrow \text{the index of label trained by } h^i_j$ \;
        $ cc_k \leftarrow cc_k+1$ \;
    	$\bm{x}' \leftarrow [x_1,...,x_d,h^i_1(x),...,h^i_{j-1}(x)]$ \;
        \If {$h^i_j(\bm{x}') = 1$}{
          $\hat{y}_k \leftarrow \hat{y}_k+1$ \;
        }
    }
  }
  \For {$j\leftarrow 1$ \KwTo $q$}{
    $\hat{y}_j \leftarrow \hat{y}_j/cc_j$ \;
  }
\KwRet{$\hat{\bm{y}} $} \;
 	
 \caption{Testing of ECCRU2}
  \label{al_ECCRU2Test}
\end{small}
\end{algorithm2e}

\subsection{Complexity Analysis}
Let's define $\Theta_{tr}(m_j,d)$ and $\Theta_{te}(d)$ the complexity of training and testing a binary classifier for label $l_j$, respectively. 
The complexity of ECCRU is $O\left(c\sum_{j=1}^{q} \Theta_{tr}(m_j,d)+ncq\Theta_{te}(d) \right)$ for training and $O\left(cq\Theta_{te}(d)\right)$ for testing.
The training and testing complexity of ECCRU2 is $O\left(\frac{c}{q}\sum_{k=1}^{q}m_k* \sum_{j=1}^{q}\left( \frac{1}{m_j}\Theta_{tr}(m_j,d)\right)+n\Theta_{te}(d) \sum_{j=1}^{q} c_j \right)$ and $O\left(\Theta_{te}(d)\sum_{j=1}^{q} c_j \right)$. For both algorithms, the first part of the training complexity concerns building classifiers and the second relates to generating the augmented feature space. 
The classifiers in ECCRU2 are more than in ECCRU, which results in larger testing complexity and large complexity of generating augmented features. However, the comparison between the training complexity of the first part of ECCRU2 and ECCRU depends on the $m_j$ and $\Theta_{tr}(m_j,d)$ of each label. The formulation of the training and testing complexity of ECCRU3 is the same with ECCUR2, but ECCRU3 is more time-consuming than ECCRU2 in both processes in practice, because a larger lower bound in the number of classifiers is applied to ECCRU3.


\section{Related Work}
A series of approaches by the same research group have been proposed for dealing with class imbalance in the context of multi-label learning using under/over-sampling. LP-RUS and LP-ROS are two twin sampling methods, of which the former removes instances assigned with most frequent labelset and the latter replicates instances whose labelset appears fewer times \citep{Charte2013}. 
ML-RUS and ML-ROS 
delete instances with majority labels and clone examples with minority labels, respectively \citep{Charte2015AddressingAlgorithms}. MLeNN is a heuristic undersampling method based on the Edited Nearest Neighbor (ENN) rule, which eliminates instances only with majority labels and similar labelset of its neighbors  \citep{Charte2014MLeNN:Undersampling}. MLSMOTE tries to make a multi-label dataset more balanced via generating synthetic instances according to a randomly selected instance containing  minority labels and its neighbors \citep{Charte2015MLSMOTE:Generation}. REMEDIAL decomposes each complex instance into two easier instances, one of which merely contains majority labels and another only with minority labels \citep{Charte2015ResamplingLabels}. 

Another kind of methods deal with the imbalance problem of multi-label learning via transforming the multi-label dataset to several binary/multi-class classification problems.
A simple strategy is dividing the multi-label dataset into several independent binary datasets, as BR does \citep{Boutell2004}, and using sampling or an ensemble strategy to solve each imbalanced binary classification problem \citep{Chen2006EfficientClassifiers,Dendamrongvit2010UndersamplingDomains,Tahir2012InverseClassification,Wan2017HPSLPred:Source}.
Cross-Coupling Aggregation (COCOA) \citep{Zhang2015TowardsLearning} is proposed to leverage the exploitation of label correlations as well as the exploration of imbalance via building one binary-class imbalance learner and several multi-class imbalance learners for each label with the assistance of sampling. The Sparse Oblique Structured Hellinger Forests (SOSHF) \citep{Daniels2017AddressingForests} transforms the multi-label learning task to an imbalanced single label classification assignment via  cost-sensitive clustering method and the transformed imbalanced classification problem is solved by tree classifiers where splitting point is determined by minimizing the sparse Hellinger loss.

In addition, some approaches that extend existing multi-label learning methods to tackle class-imbalance problem have been proposed, such as neural network based \citep{Tepvorachai2008Multi-labelTraining,Li2013ImprovementSamples,Sozykin2017}, SVM based \citep{Cao2016} and hypernetwork based \citep{Sun2017AddressingHypernetwork}. Finally, other strategies, such as representation learning \citep{LiWang16}, constrained submodular minimization \citep{Wu:2016} and balanced pseudo-label \citep{Zeng2014}, have been utilized to address the imbalance problem of multi-label learning as well.

Compared to the above approaches, the strengths of the proposed methods are as follows. Firstly, they build on top of a theoretically grounded and highly accurate method, ECC. Secondly, they inherit the ability of ECC to model correlation among many labels, in contrast for example to \citep{Zhang2015TowardsLearning} that is second-order and \citep{Chen2006EfficientClassifiers,Dendamrongvit2010UndersamplingDomains,Tahir2012InverseClassification,Wan2017HPSLPred:Source} that are first-order methods. Thirdly, it is algorithm independent, as it can be combined with any binary classifier that best fits the problem at hand, in contrast to \citep{Tepvorachai2008Multi-labelTraining,Li2013ImprovementSamples,Sozykin2017,Cao2016,Sun2017AddressingHypernetwork,Daniels2017AddressingForests} that build on top of particular learning paradigms.  

\section{Empirical Analysis}
We first introduce the setup of our experiments. Then we present the experimental results and their analysis. 

\subsection{Setup}
Our empirical study is based on 16 multi-label data sets obtained from Mulan's GitHub repository\footnote{\href{https://github.com/tsoumakas/mulan/tree/master/data/multi-label}{https://github.com/tsoumakas/mulan/tree/master/data/multi-label}} \citep{Tsoumakas2011MULAN:Learning}. 
Table \ref{ta_dataset} lists these datasets along with their main statistics. In textual data sets with more than 1000 features we applied a simple dimensionality reduction approach that retains the top 10$\%$ (bibtex, enron, eurlex-sm, medical) or top 1$\%$ (rcv1subset1, rcv1subset2, yahoo-Arts1, yahoo-Business1) of the features ordered by number of non-zero values (i.e. frequency of appearance), similar to \citep{Zhang2015TowardsLearning}. 

The proposed approaches are compared against five multi-label learning methods. Two of them are imbalance agnostic ones, namely the Binary Relevance (BR) baseline \citep{Boutell2004} and the state-of-the-art ECC \citep{Read2011ClassifierClassification}, on which the proposed approaches build on. The other three methods are imbalance aware ones that similarly to ours are based on random undersampling, namely BR with random undersampling (BRUS), ensemble of BRUS (EBRUS) and the state-of-the-art COCOA \citep{Zhang2015TowardsLearning}. In ECCRU2 and ECCRU3, the $\theta_{max}$ is set to 10. The $\theta_{min}$ in ECCRU3 is set to 0.5. The ensemble size is set to 10 for all ensemble methods (ECC, EBRUS, ECCRU*). For COCOA in particular, the number of coupling class labels is set to $min(q-1,10)$ as in \citep{Zhang2015TowardsLearning}. A decision tree is used as the base classifier in all methods. 

We employ five widely used binary metrics for imbalanced data \citep{He2009LearningData,Akosa2017PredictiveNegative}: F-measure, G-mean, Balanced Accuracy, area under the receiver operating characteristic curve (AUC-ROC) and area under the precision-recall curve (AUC-PR). The first three are computed on top of binary predictions, while the last two on top of ranked lists of test instances in order of relevance to the positive class, which most of the times is the minority class. For all  these metrics, the higher their value, the better the accuracy of the corresponding algorithm. Binary predictions are obtained after setting a separate threshold $t \in \{0, 0.05, \ldots, 1\}$ per label. This threshold is set so as to maximize the corresponding evaluation metric (F-measure, G-mean, Balanced Accuracy) on the training set. 


We compute the average of the above metrics across all labels, an approach to aggregating binary measures in multi-class and multi-label tasks that is called {\em macro-averaging}. The alternative approach, {\em micro-averaging}, collects the predictions for all labels as if they were part of a single binary classification task. Macro-averaging is more suitable for imbalanced learning as it treats all labels equally, in contrast to micro-averaging where the contribution of each label depends on the frequency of the positive class \citep{Tang2009LargeMetalabeler}. We apply $5 \times 2$-fold cross validation with multi-label stratification \citep{Sechidis2011} to each dataset and the average results are reported. 

We obtained the code of the five existing methods from Mulan. Our approaches were also implemented in the context of Mulan. The experiments were conducted on a machine with 4 10-core CPUs running at 2.27 GHz.  


\begin{table}[t]
\centering
\small
\caption{The 16 multi-label data sets used in this study. Columns $n$, $d$, $q$ denote the number of instances, features and labels respectively, $LC$ the label cardinality, $MeanImR$ and $MaxImR$ the average and maximum $ImR$ of the labels and $CVImR$ the normalized standard deviation of the $ImR$ of the labels.}
\label{ta_dataset}
\begin{tabular}{@{}ccccccccc@{}}
\toprule
DataSet         & Domain  & \textit{n} & \textit{d} & \textit{q} & \textit{LC} & $MeanImR$  & $MaxImR$   & $CVImR$ \\ \midrule
bibtex & text & 7395 & 183 & 159 & 2.402 & 87.699 & 144 & 0.410 \\
cal500 & music & 502 & 68 & 174 & 26.044 & 22.345 & 99.4 & 1.129 \\
corel5k & image & 5000 & 499 & 374 & 3.522 & 845.284 & 4999 & 1.528 \\
enron & text & 1702 & 100 & 53 & 3.378 & 136.867 & 1701 & 1.974 \\
eurlex-sm & text & 19348 & 500 & 201 & 2.213 & 2420.775 & 19347 & 2.136 \\
flags & image & 194 & 19 & 7 & 3.392 & 2.753 & 6.462 & 0.711 \\
genbase & biology & 662 & 1186 & 27 & 1.252 & 143.458 & 661 & 1.460 \\
mediamill & video & 43907 & 120 & 101 & 4.376 & 331.439 & 1415.355 & 1.178 \\
medical & text & 978 & 144 & 45 & 1.245 & 328.069 & 977 & 1.151 \\
rcv1subset1 & text & 6000 & 472 & 101 & 2.88 & 235.58 & 2999 & 2.089 \\
rcv1subset2 & text & 6000 & 472 & 101 & 2.634 & 190.906 & 1999 & 1.724 \\
scene & image & 2407 & 144 & 6 & 1.074 & 4.662 & 5.613 & 0.148 \\
tmc2007-500 & text & 28596 & 500 & 22 & 2.158 & 25.823 & 63.844 & 0.791 \\
yahoo-Arts1 & text & 7484 & 231 & 26 & 1.548 & 384.756 & 7483 & 3.816 \\
yahoo-Business1 & text & 11214 & 219 & 30 & 1.437 & 1014.363 & 11213 & 2.813 \\
yeast & biology & 2417 & 103 & 14 & 4.237 & 8.954 & 70.088 & 1.997 \\ \bottomrule
\end{tabular}
\end{table}

\subsection{Results and Analysis} 

We discuss the experimental results from two aspects. We first report the accuracy and  training time of all participating methods over the 16 multi-label datasets and present results of significance tests. We then discuss how different methods behave under different levels of imbalance ratio. 

\begin{table}
\centering
\small
\caption{Average rank of all methods in terms of five evaluation metrics and training time.}
\label{ta_aveRank}
\begin{tabular}{@{}rcccccccc@{}}
\toprule
     & BR   & ECC  & BRUS & EBRUS & COCOA & ECCRU & ECCRU2 & ECCRU3 \\ \midrule
F-measure & 6.44 & 3.09 & 6.56 & 5.31  & 4.22  & 4.34  & 3.22    & \textbf{2.81}     \\
G-mean & 7.75 & 5.81 & 3.31 & 5.63  & 5.00     & \textbf{2.50}   & 3.28    & 2.72     \\
Balanced Acc. & 7.63 & 5.38 & 6.25 & 5.38  & 4.50   & 2.38  & 2.59    & \textbf{1.91}     \\
AUC-ROC & 7.81 & 5.50  & 6.88 & 3.81  & 4.31  & 3.00     & 2.84    & \textbf{1.84}     \\
AUC-PR  & 7.00    & 2.63 & 7.88 & 5.06  & 4.50   & 3.28  & 3.59    & \textbf{2.06}     \\
Training Time & 2.88 & 7.69 & \textbf{1.00} & 5.06  & 3.19  & 5.38  & 5.00    & 5.81     \\ \bottomrule
\end{tabular}
\end{table}

Table \ref{ta_aveRank} shows the average rank of each method in terms of the five evaluation metrics plus the training time, with the best result highlighted with bold typeface. 
We first notice that ECCRU3 achieves the best results in all evaluation metrics, with the exception of G-mean, where it is second best behind ECCRU. In addition, the proposed methods achieve the top 3 positions for all evaluation metrics, with the exception of F-measure and AUC-PR where ECC achieves the second position. In terms of training time, BRUS achieves the best results followed by BR and COCOA. The proposed methods come next, followed by ECC that achieves the worst results.      
To examine the statistical significance of the differences between the different methods participating in our empirical study, we employ the Friedman test, followed by the Wilcoxon signed rank test with Bergman-Hommel's correction at the 5\% level, following literature guidelines \citep{Garcia2008AnComparisons,Benavoli2016ShouldMean-Ranks}. Table \ref{ta_WtestSummary} presents the results. We notice that ECCRU3 achieves the most significant wins than any other method in all measures: 4 in F-measure, 4 in G-mean together with BRUS and the other two variations of the proposed approach, 6 in Balanced Accuracy, 7 in AUC-ROC, and 6 in AUC-PR together with ECC. BRUS in G-mean, COCOA and ECC in F-measure and ECC in AUC-PR are the only 4 out of 25 cases of non-significant difference between ECCRU3 and the five competing methods in the five evaluation measures. 
In terms of training time, BRUS has the most wins minus losses (7), followed by COCOA (4), BR (3), EBRUS and ECCRU2 (-1), ECCRU (-2), ECCRU3 (-3) and ECC (-7). While ECC is competitive with ECCRU3 in F-measure and AUC-PR, it is significantly worse in training time. If G-mean (F-measure) is the measure of interest, then BRUS (COCOA) is an algorithm to consider as it is both highly accurate and efficient.     

\begin{table}[htb]
\centering
\small
\caption{Results of the Wilcoxon signed rank test with Bergman-Hommel's correction at the 5\% level among all pairs of methods. ``$\uparrow$" (``$\downarrow$") denotes the method in bold typeface in the upper-left corner of each subtable is significantly superior (inferior) to the corresponding method of each row. ``-" denotes lack of significant difference between the two methods. Abbreviations stand for: (F)-measure, (G)-mean, (B)alanced Accuracy, AUC-(R)OC, AUC-(P)R, Training (T)ime.}
\label{ta_WtestSummary}
\begin{tabular}{lcccccc|lcccccc}
\toprule
\textbf{BR vs} & \textit{F} & \textit{G} & \textit{B} & \textit{R} & \textit{P} & \textit{T} & \textbf{ECC vs} & \textit{F} & \textit{G} & \textit{B} & \textit{R} & \textit{P} & \textit{T} \\ \midrule
ECC & $\downarrow$ & $\downarrow$ & $\downarrow$ & $\downarrow$ & $\downarrow$ & $\uparrow$ & BR & $\uparrow$ & $\uparrow$ & $\uparrow$ & $\uparrow$ & $\uparrow$ & $\downarrow$ \\
BRUS & - & $\downarrow$ & $\downarrow$ & $\downarrow$ & $\uparrow$ & $\downarrow$ & BRUS & $\uparrow$ & $\downarrow$ & - & $\uparrow$ & $\uparrow$ & $\downarrow$ \\
EBRUS & - & $\downarrow$ & $\downarrow$ & $\downarrow$ & $\downarrow$ & - & EBRUS & $\uparrow$ & - & - & $\downarrow$ & $\uparrow$ & $\downarrow$ \\
COCOA & - & $\downarrow$ & $\downarrow$ & $\downarrow$ & $\downarrow$ & - & COCOA & - & - & - & $\downarrow$ & $\uparrow$ & $\downarrow$ \\
ECCRU & - & $\downarrow$ & $\downarrow$ & $\downarrow$ & $\downarrow$ & $\uparrow$ & ECCRU & - & $\downarrow$ & $\downarrow$ & $\downarrow$ & $\uparrow$ & $\downarrow$ \\
ECCRU2 & $\downarrow$ & $\downarrow$ & $\downarrow$ & $\downarrow$ & $\downarrow$ & $\uparrow$ & ECCRU2 & - & $\downarrow$ & $\downarrow$ & $\downarrow$ & $\uparrow$ & $\downarrow$ \\
ECCRU3 & $\downarrow$ & $\downarrow$ & $\downarrow$ & $\downarrow$ & $\downarrow$ & $\uparrow$ & ECCRU3 & - & $\downarrow$ & $\downarrow$ & $\downarrow$ & - & $\downarrow$ \\
\toprule
\textbf{BRUS vs} & \textit{F} & \textit{G} & \textit{B} & \textit{R} & \textit{P} & \textit{T} & \textbf{EBRUS vs} & \textit{F} & \textit{G} & \textit{B} & \textit{R} & \textit{P} & \textit{T} \\
\hline
BR & - & $\uparrow$ & $\uparrow$ & $\uparrow$ & $\downarrow$ & $\uparrow$ & BR & - & $\uparrow$ & $\uparrow$ & $\uparrow$ & $\uparrow$ & - \\
ECC & $\downarrow$ & $\uparrow$ & - & $\downarrow$ & $\downarrow$ & $\uparrow$ & ECC & $\downarrow$ & - & - & $\uparrow$ & $\downarrow$ & $\uparrow$ \\
EBRUS & - & $\uparrow$ & - & $\downarrow$ & $\downarrow$ & $\uparrow$ & BRUS & - & $\downarrow$ & - & $\uparrow$ & $\uparrow$ & $\downarrow$ \\
COCOA & - & $\uparrow$ & $\downarrow$ & $\downarrow$ & $\downarrow$ & $\uparrow$ & COCOA & - & $\downarrow$ & - & - & - & $\downarrow$ \\
ECCRU & - & - & $\downarrow$ & $\downarrow$ & $\downarrow$ & $\uparrow$ & ECCRU & - & $\downarrow$ & $\downarrow$ & - & $\downarrow$ & - \\
ECCRU2 & $\downarrow$ & - & $\downarrow$ & $\downarrow$ & $\downarrow$ & $\uparrow$ & ECCRU2 & $\downarrow$ & $\downarrow$ & $\downarrow$ & - & $\downarrow$ & - \\
ECCRU3 & $\downarrow$ & - & $\downarrow$ & $\downarrow$ & $\downarrow$ & $\uparrow$ & ECCRU3 & $\downarrow$ & $\downarrow$ & $\downarrow$ & $\downarrow$ & $\downarrow$ & - \\
\toprule
\textbf{COCOA vs} & \textit{F} & \textit{G} & \textit{B} & \textit{R} & \textit{P} & \textit{T} & \textbf{ECCRU vs} & \textit{F} & \textit{G} & \textit{B} & \textit{R} & \textit{P} & \textit{T} \\
\hline
BR & - & $\uparrow$ & $\uparrow$ & $\uparrow$ & $\uparrow$ & - & BR & - & $\uparrow$ & $\uparrow$ & $\uparrow$ & $\uparrow$ & $\downarrow$ \\
ECC & - & - & - & $\uparrow$ & $\downarrow$ & $\uparrow$ & ECC & - & $\uparrow$ & $\uparrow$ & $\uparrow$ & $\downarrow$ & $\uparrow$ \\
BRUS & - & $\downarrow$ & $\uparrow$ & $\uparrow$ & $\uparrow$ & $\downarrow$ & BRUS & - & - & $\uparrow$ & $\uparrow$ & $\uparrow$ & $\downarrow$ \\
EBRUS & - & $\uparrow$ & - & - & - & $\uparrow$ & EBRUS & - & $\uparrow$ & $\uparrow$ & - & $\uparrow$ & - \\
ECCRU & - & $\downarrow$ & $\downarrow$ & $\downarrow$ & - & $\uparrow$ & COCOA & - & $\uparrow$ & $\uparrow$ & $\uparrow$ & - & $\downarrow$ \\
ECCRU2 & - & $\downarrow$ & $\downarrow$ & $\downarrow$ & - & $\uparrow$ & ECCRU2 & - & - & - & - & - & - \\
ECCRU3 & - & $\downarrow$ & $\downarrow$ & $\downarrow$ & $\downarrow$ & $\uparrow$ & ECCRU3 & $\downarrow$ & - & - & $\downarrow$ & $\downarrow$ & - \\
\toprule
\textbf{ECCRU2 vs} & \textit{F} & \textit{G} & \textit{B} & \textit{R} & \textit{P} & \textit{T} & \textbf{ECCRU3 vs} & \textit{F} & \textit{G} & \textit{B} & \textit{R} & \textit{P} & \textit{T} \\
\hline
BR & $\uparrow$ & $\uparrow$ & $\uparrow$ & $\uparrow$ & $\uparrow$ & $\downarrow$ & BR & $\uparrow$ & $\uparrow$ & $\uparrow$ & $\uparrow$ & $\uparrow$ & $\downarrow$ \\
ECC & - & $\uparrow$ & $\uparrow$ & $\uparrow$ & $\downarrow$ & $\uparrow$ & ECC & - & $\uparrow$ & $\uparrow$ & $\uparrow$ & - & $\uparrow$ \\
BRUS & $\uparrow$ & - & $\uparrow$ & $\uparrow$ & $\uparrow$ & $\downarrow$ & BRUS & $\uparrow$ & - & $\uparrow$ & $\uparrow$ & $\uparrow$ & $\downarrow$ \\
EBRUS & $\uparrow$ & $\uparrow$ & $\uparrow$ & - & $\uparrow$ & - & EBRUS & $\uparrow$ & $\uparrow$ & $\uparrow$ & $\uparrow$ & $\uparrow$ & - \\
COCOA & - & $\uparrow$ & $\uparrow$ & $\uparrow$ & - & $\downarrow$ & COCOA & - & $\uparrow$ & $\uparrow$ & $\uparrow$ & $\uparrow$ & $\downarrow$ \\
ECCRU & - & - & - & - & - & - & ECCRU & $\uparrow$ & - & - & $\uparrow$ & $\uparrow$ & - \\
ECCRU3 & - & - & $\downarrow$ & $\downarrow$ & $\downarrow$ & $\uparrow$ & ECCRU2 & - & - & $\uparrow$ & $\uparrow$ & $\uparrow$ & $\downarrow$ \\
\bottomrule
\end{tabular}
\end{table}

\begin{figure}[ht]
\begin{center}
\small
\subfigure[Percentage of labels]{%
\includegraphics[width=.45\textwidth]{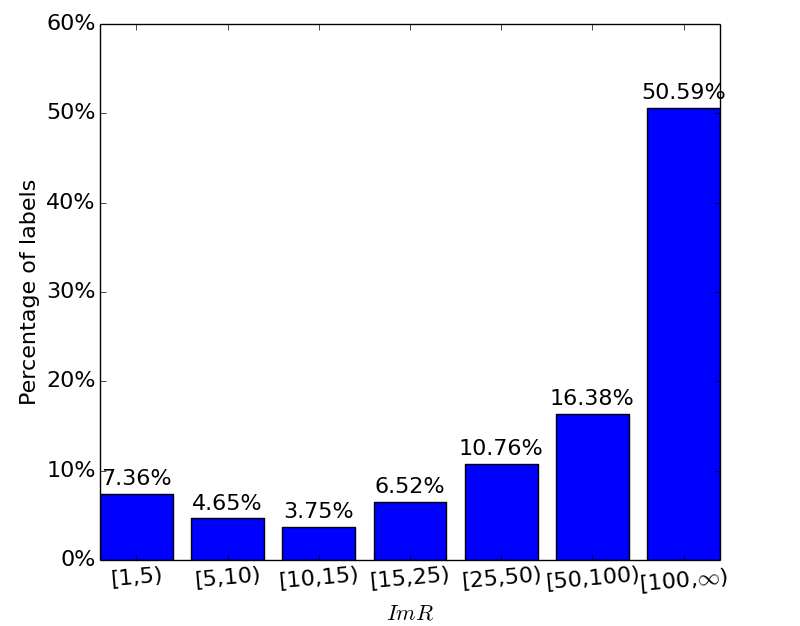}} \label{fig_Per}
\subfigure[F-measure]{%
\includegraphics[width=.45\textwidth]{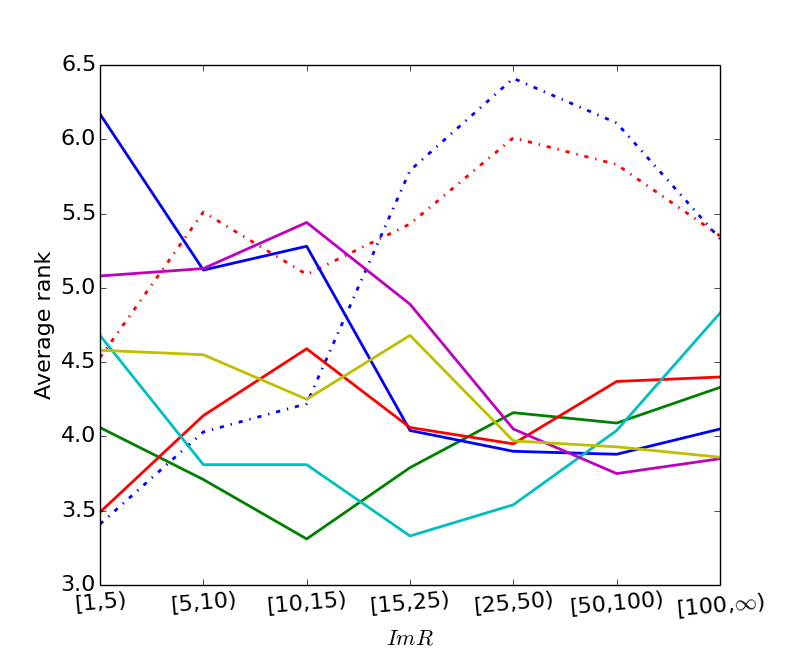}} \label{fig_F}\
\subfigure[G-mean]{%
\includegraphics[width=.45\textwidth]{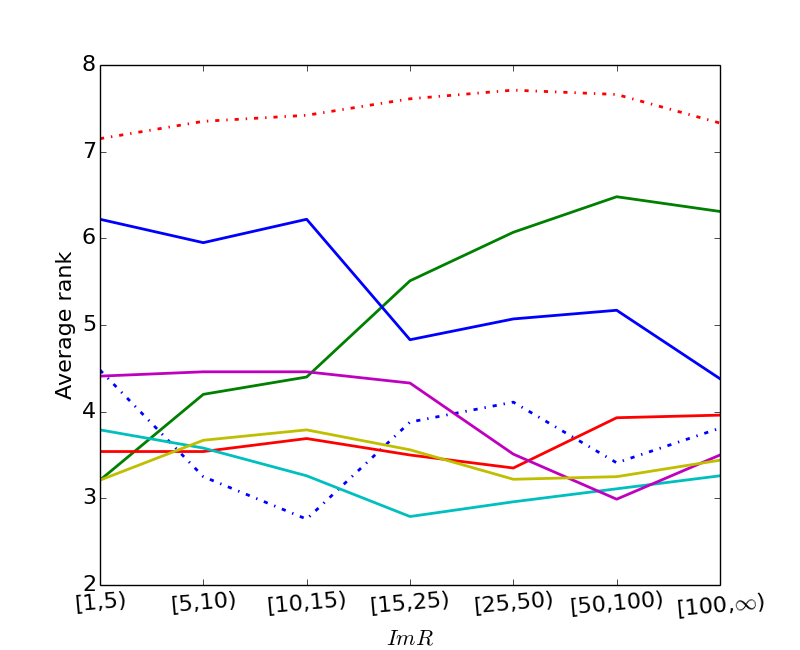}} \label{fig_G} 
\subfigure[Balanced Accuracy]{%
\includegraphics[width=.45\textwidth]{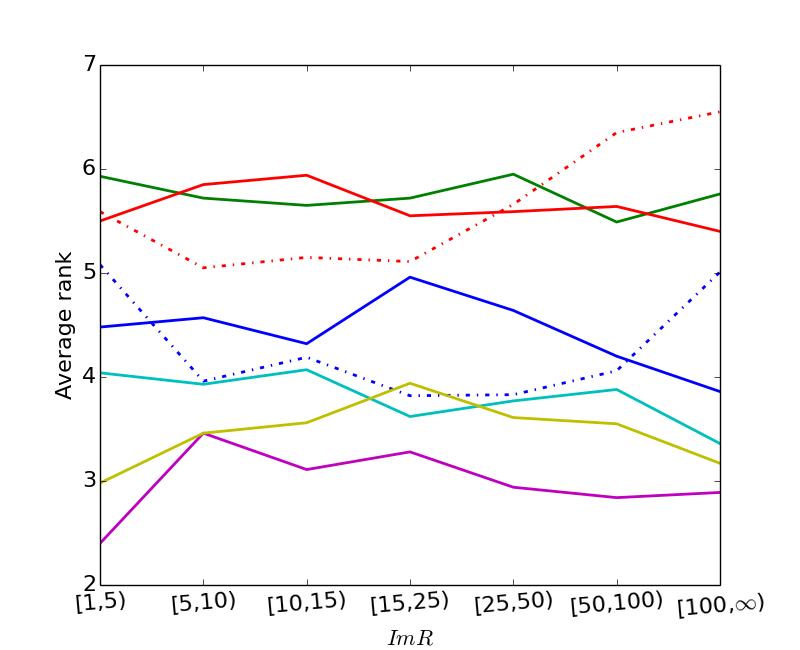}} \label{fig_B} \
\subfigure[AUC-ROC]{%
\includegraphics[width=.45\textwidth]{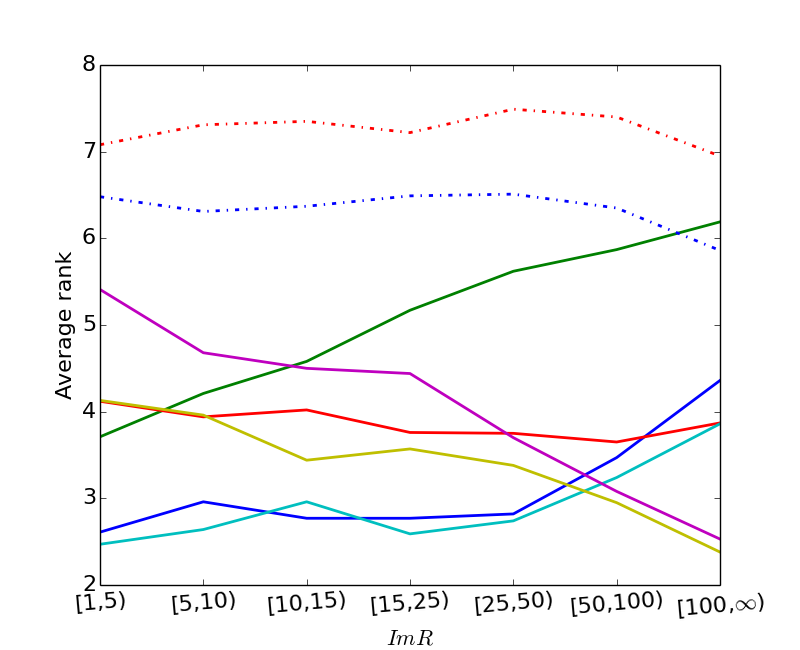}} \label{fig_A}
\subfigure[AUC-PR]{%
\includegraphics[width=.45\textwidth]{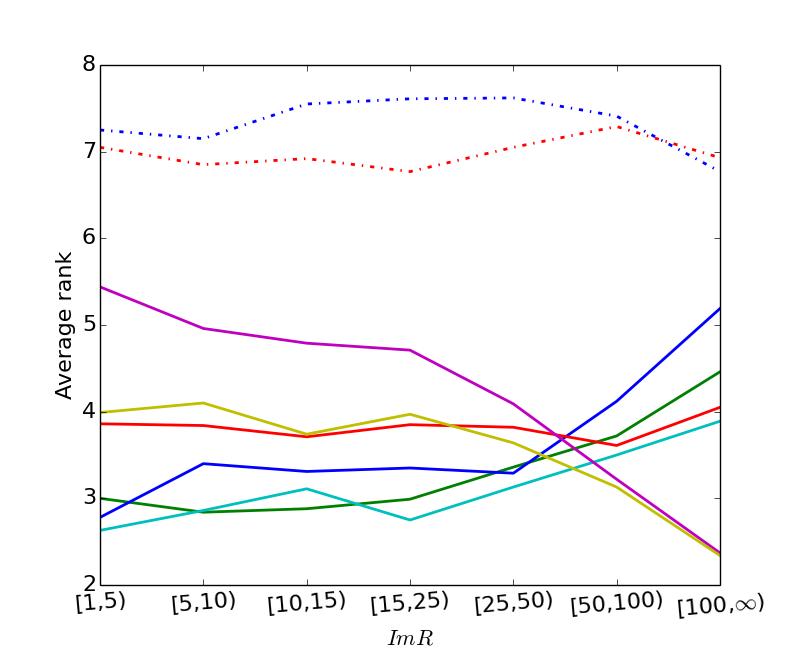}} \label{fig_P} \

\begin{minipage}{0.9\linewidth}
\centerline{\includegraphics[width=1\textwidth]{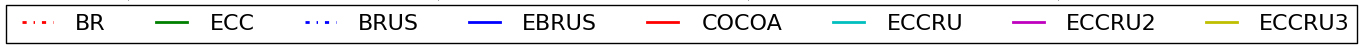}}
\end{minipage}

\caption{Sub-figure (a) shows the percentage of labels and sub-figures (b)-(f) the average rank of all methods for different $ImR$ intervals in terms of the 5 different evaluation metrics.}\label{fig_variousImR}
\end{center}
\end{figure}

To investigate the accuracy of the competing methods under different imbalance levels, we divide $ImR$ into 7 intervals: $[1,5)$, $[5,10)$, $[10,15)$, $[15,25)$, $[25,50)$, $[50,100)$, $[100,\infty)$. Figure \ref{fig_variousImR} (a) shows the percentages of the labels from all 16 datasets that fall into these intervals. We can see that more than half of the labels have $ImR \geq 100$ and only 16\% of the labels have $ImR < 15$. Figure \ref{fig_variousImR} (b)-(f) shows the average rank of the 8 competing methods based on the 5 evaluation metrics calculated on each subset of all labels belonging to each interval. We can see that the proposed approaches dominate the rankings for all measures when $ImR \geq 15$. When $ImR < 15$ COCOA and BRUS do well in F-measure and G-mean, EBRUS does well in AUC-ROC and AUC-PR and ECC does well in F-measure, G-mean and AUC-PR. 

We also notice that with the exception of Balanced Accuracy, ECCRU3 dominates ECCRU2 in all other measures for roughly all levels of $ImR$, and ECCRU dominates ECCRU2 and ECCRU3 for $ImR < 50$. This hints that a meta-approach selecting ECCRU for low $ImR$ and ECCRU3 for high $ImR$ could lead to even better results. We hypothesize that the observed behavior is due to the following reason: when imbalance is not that large, then it is more important to build the full chains of ECCRU, in order to gain from modeling the dependencies among the labels, similarly to ECC. When imbalance is large, then it is more important to direct training effort towards exploiting more of the majority samples in imbalanced labels that are part of smaller chains. In other words, given the same budget of training examples, when $ImR$ is high, then the benefits from exploiting more majority training examples surpass the benefits of modeling label dependencies.

\section{Conclusion}
We started from a strong and theoretically grounded multi-label learning algorithm, ECC, and made it resilient to the challenge of class imbalance by employing random undersampling to balance the class distribution of each binary training set, leading to the ECCRU algorithm. We then discussed approaches to make the best exploitation of a computational budget based on the key observation that different imbalance ratios lead to different levels of exploitation of majority examples, leading to the ECCRU2 and ECCRU3 algorithms. Our empirical study showed that the proposed method are competitive to related benchmark and state-of-the-art methods, and especially ECCRU3 achieves the best performance in terms of five imbalance metrics with positive significance tests in almost all comparisons. We also presented an interesting analysis of the behavior of the algorithms under different levels of class imbalance, and discussed insights on the causes of this behavior. 

\acks{Bin Liu is supported from the China Scholarship Council (CSC) under the Grant CSC No.201708500095. Grigorios Tsoumakas is partially supported from Atypon Systems LLC.}

\bibliography{Mendeley_alma.bib}

\end{document}